# SCHIGAND: A Synthetic Facial Generation Model Pipeline

*Ananya Kadali, Sunnie Jehan-Morrison, Orasiki Wellington, Barney Evans, Precious Durojaiye, Richard Guest**

*School of Electronics and Computer Science, University of Southampton, Southampton, United Kingdom*
**R.M.Guest@soton.ac.uk*

## Abstract

The growing demand for diverse and high-quality facial datasets for training and testing biometric systems is challenged by privacy regulations, data scarcity, and ethical concerns. Synthetic facial images offer a potential solution, yet existing generative models often struggle to balance realism, diversity, and identity preservation. This paper presents SCHIGAND, a novel synthetic face generation pipeline integrating StyleCLIP, HyperStyle, InterfaceGAN, and Diffusion models to produce highly realistic and controllable facial datasets. SCHIGAND enhances identity preservation while generating realistic intra-class variations and maintaining inter-class distinctiveness, making it suitable for biometric testing. The generated datasets were evaluated using ArcFace, a leading facial verification model, to assess their effectiveness in comparison to real-world facial datasets. Experimental results demonstrate that SCHIGAND achieves a balance between image quality and diversity, addressing key limitations of prior generative models. This research highlights the potential of SCHIGAND to supplement and, in some cases, replace real data for facial biometric applications, paving the way for privacy-compliant and scalable solutions in synthetic dataset generation.

## 1  Introduction

Biometric systems utilise personal characteristics to authenticate or identify individuals. These characteristics can be categorised as either behavioural traits or physical attributes. State-of-the-art systems are typically deep learning models that require substantial amounts of data for training and testing, often comprising real facial images. However, collecting large-scale and diverse facial datasets across various poses and environments introduces significant challenges. These challenges include high costs, considerable time investment, privacy concerns, restricted access due to strict compliance with legislation such as the General Data Protection Regulation (GDPR) [1], the lack of anonymisation in some datasets, and other related issues. Recent advancements in generative artificial intelligence (GenAI) have introduced synthetic facial imagery as a viable solution to these issues. Synthetic data enables the creation of realistic and diverse datasets suitable for biometric testing while addressing privacy concerns, reducing bias, and potentially lowering resource expenditure.

This paper investigates the potential of synthetic facial imagery, generated using advanced GenAI models, to evaluate the capabilities of biometric systems in facial recognition. A key aspect of the project involves implementing a novel pipeline as described in Section 3 comprising GenAI models to create synthetic images. The paper focuses on assessing the feasibility and effectiveness of synthetic data in scenarios where obtaining large, authentic and demographically representative datasets proves challenging.

## 2  Literature Review

The development of synthetic facial datasets has gained significant attention in recent years, particularly for applications in face recognition and identification. Various generative models have been proposed to create realistic synthetic faces, each addressing specific challenges such as identity preservation, intra-class variation, and dataset diversity. The following models represent key advancements in synthetic face generation and serve as benchmarks for comparison with our proposed pipeline, SCHIGAND.

DCFace is a synthetic face generation framework proposed by Kim et al. [2] designed to produce high-quality facial images for training and evaluating facial recognition systems. It addresses key challenges in synthetic data generation by aiming to balance identity preservation with diversity specifically, generating synthetic images that reflect both inter-class and intra-class variations. It employs a two-stage generative framework to produce synthetic images. In the first stage, an unconditional denoising diffusion probabilistic model (DDPM), trained on the Flickr-FaceHQ (FFHQ) [3] dataset, generates synthetic identity subjects. From this set of generated subjects, one identity subject is selected for further processing. In the second stage, a "style bank" composed of real-face data that determines the final image's overall appearance and stylistic characteristics is utilised. This style bank, along with the selected identity subject, is given as input to a conditional DDPM trained on the CASIA-WebFace dataset [4][5]. During this stage, a mixing process takes place, resulting in diverse synthetic images that retain the identity traits of the selected identity subject while incorporating the stylistic features of the style bank [2]. The efficacy of DCFace is assessed by evaluating the generated images using pre-trained facial recognition models, and it has been shown to yield both inter-class and intra-class variations in synthetic subjects. In the 0.5M image training setting, DCFace outperformed methods, SynFace and DigiFace, on four out of five face verification benchmarks (LFW, CPLFW, AgeDB, CALFW), achieving an average verification accuracy improvement of 6.11% over these methods [2]. DCFace offers strong identity preservation and intra-class and inter-class diversity through dual conditioning but lacks the fine-grained attribute-level control such as expression, pose, or accessories that alternative methods like Generative Adversarial Network (GAN) based models offer. An aim of this study is to address this limitation by enhancing identity retention while ensuring greater diversity and attribute control.

GANDiffFace [6] improves image generation by combining the strengths of Generative Adversarial Networks (GANs) and diffusion models, addressing the limited variation of GANs and the lower image quality of diffusion models. GANs are limited by poor intra-class variation, struggling to produce diverse depictions of a single identity. Conversely, while diffusion models excel at generating this variation, they often do so at the cost of realism compared to GAN-based approaches. GANDiffFace's ability to synthesise realistic faces with high intra-class variation is therefore highly relevant to our research, as it can effectively simulate the diversity of faces found in real-world, uncontrolled datasets. The GANDiffFace architecture, which enables this combination, is illustrated in the pipeline in Figure 1. The GAN model (StyleGAN3 [7]) generates a synthetic identity with controllable attributes, such as race, gender, and age, and produces simple variations of that identity. These variations are then passed to the DreamBooth module [6] which combines this data with a list of prompts to create the diffusion output. Although GANDiffFace improves diversity, it does so at the expense of identity preservation, as its focus is on maximising intra-class variations rather than maintaining a consistent identity across variations. A key objective of the current research is to address this limitation by introducing mechanisms that retain identity consistency while still allowing controlled variation, making it better suited for biometric dataset generation.

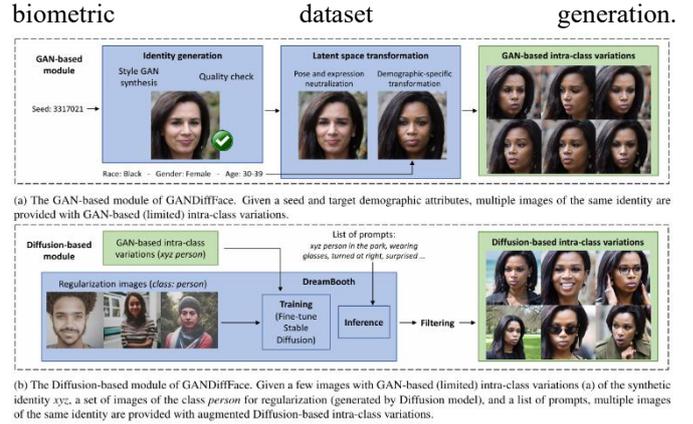

(a) The GAN-based module of GANDiffFace. Given a seed and target demographic attributes, multiple images of the same identity are provided with GAN-based (limited) intra-class variations.

(b) The Diffusion-based module of GANDiffFace. Given a few images with GAN-based (limited) intra-class variations (a) of the synthetic identity *id_i*, a set of images of the class *person* for regularization (to generate by Diffusion models), and a list of prompts, multiple images of the same identity are provided with augmented Diffusion-based intra-class variations.

*Figure 1. GAN and Diffusion based modules of the GANDiffFace Architecture.*

Stable Diffusion is a text-to-image generation model introduced 2022 by Stability AI [8]. It employs a latent diffusion model (LDM) that uses Contrastive Language-Image Pre-training (CLIP) to map text prompts into a noisy latent image space, which is then de-noised to create coherent visuals [8]. A key feature of stable diffusion is its ability to preserve identity consistency through seed values, ensuring reproducibility across generated outputs. It is a recent innovation from the creators of Stable Diffusion which integrates a hybrid architecture of multimodal and parallel diffusion transformers. It surpasses predecessors and competitors like DALL-E and MidJourney in output diversity and scene complexity [9]. Despite its superior performance across metrics such as style and coherence, ethical concerns regarding the datasets potentially used by Stability AI (Stable Diffusion), which may rely on uncurated data collection from the Internet, have raised questions about its suitability for facial biometrics [8]. Efforts to address these concerns include adding safety filters to Stable Diffusion models [11] and cleaning training datasets in later versions [12]. An aim of this current research is to improve upon the limitations of Stable Diffusion by incorporating structures text-prompt engineering specifically tailored for facial biometrics testing, as opposed to generalised image generation.

# 3    Components of SCHIGAND

SCHIGAND is an acronym for StyleClip Hyperstyle InterfaceGAN Diffusion. In this section, we will discuss the different components of this pipeline and their importance.

Figure 2 provides an overall visualisation of the SCHIGAND pipeline, illustrating the flow of data and the interactions between its various components. This figure also highlights key intermediate outputs generated at each stage, offering a comprehensive understanding of how the model progresses from initial input to the final generated images.

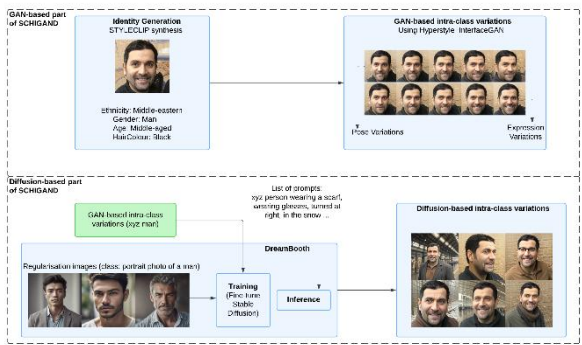

*Figure 2. Overall Architecture of the SCHIGAND Pipeline, detailing data flow and intermediate outputs.*

### 3.1 StyleCLIP (SC)

At the core of SCHIGAND lies StyleCLIP (StyleGAN + CLIP), an innovative integration of GANs with language-image alignment techniques. This combination enables the pipeline to synthesise facial images directly from textual descriptions, providing an innovative approach to text-driven identity creation. NVIDIA's StyleGAN3 [13] was initially chosen for its advanced capabilities, including architectural improvements that minimised aliasing and ensured smoother transformations, making it a strong foundation for this component of SCHIGAND [13]

To support realistic and diverse image generation, the pipeline utilised a pre-trained StyleGAN model trained on the FFHQ dataset, introduced by Karras et al [3]. This diversity, coupled with variations in lighting, expressions, and backgrounds, provided a robust training foundation. Leveraging this pre-trained model accelerated SCHIGAND's development and ensured the generated outputs captured both semantic richness and real-world diversity.

StyleCLIP's implementation relies on the synergy between StyleGAN3 and OpenAI's CLIP model. StyleGAN3 served as the generative backbone, while CLIP iteratively refined candidate images by aligning them with textual prompts using its text-image scoring mechanism. This process allowed for precise translation of descriptive inputs into realistic visual outputs. CLIP's ability to learn visual concepts from natural language supervision, as demonstrated by Radford et al., played a pivotal role in enabling this functionality [14].

To further enhance text-image alignment, the Vision Transformer model ViT-B/16 [15] was selected as the CLIP backbone due to its fine-grained resolution and superior attention to detail. Compared to ViT-B/32, the smaller patch size of ViT-B/16 allowed for more accurate rendering of subtle features, such as facial textures and structures. These attributes were instrumental in producing high-quality outputs that adhered closely to textual prompts while maintaining visual realism.

Despite its strengths, the initial StyleCLIP implementation faced challenges in capturing certain demographic traits, particularly hair textures (as observed in Figure 3 and Figure 4) and other nuanced features, which can be attributed to biases in the training datasets. The difficulty in generating detailed hair textures across demographics highlighted the ongoing need for more inclusive and representative training data. Additionally, while increasing iteration steps improved alignment with textual descriptions, it often came at the cost of image realism, necessitating a balance capped at 20 iterations to achieve optimal results. Examples of StyleCLIP outputs are illustrated in Figure 3 and Figure 4.

As SCHIGAND evolved, challenges also emerged when integrating InterfaceGAN for pose manipulation. StyleGAN3's latent space, optimised for alias-free synthesis and geometric consistency, proved less compatible with the structured and linear manipulations required by InterfaceGAN. To address this, StyleGAN2 was adopted for its more interpretable and linear latent space, made possible by advancements such as restructured normalisation and improved path length regularisation [16]. This transition significantly expanded SCHIGAND's capabilities, enabling the generation of synthetic identities with adjustable poses while preserving both realism and diversity.

| Race | Age 4 | Age 8 | Age 12 | Age 35 | Age 75 |
|------|-------|-------|--------|--------|--------|
| **Indian** | | | | | |
| **Black** | | | | | |
| **Asian** | | | | | |
| **No-Race** | | | | | |

*Figure 3. StyleCLIP results using StyleGAN3 with standardised prompts (e.g., "A <age>-year-old <race/no race> male"), showing output diversity across ages (4, 8, 12, 35, 75) and ethnicities (Indian, Black, Asian, No-Race).*

| Race | Age 4 | Age 8 | Age 12 | Age 35 | Age 75 |
|---|---|---|---|---|---|
| Indian | 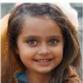 | 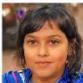 | 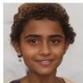 | 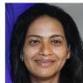 | 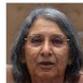 |
| Black | 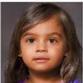 | 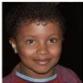 | 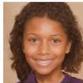 | 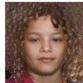 | 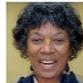 |
| Asian | 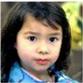 | 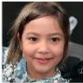 | 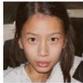 | 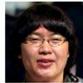 | 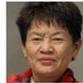 |
| No-Race | 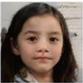 | 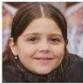 | 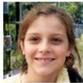 | 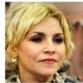 | 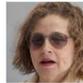 |

*Figure 4. StyleCLIP results using StyleGAN3 with standardised prompts (e.g., "A <age>-year-old <race/no race> female"), showing output diversity across ages (4, 8, 12, 35,75) and ethnicities (Indian, Black, Asian, No-Race).*

### 3.2 Hyperstyle + InterfaceGAN (HIGAN)

GANs, particularly the StyleGAN family, have transformed image synthesis by enabling high-quality visuals with semantically rich latent spaces. These latent spaces support diverse editing applications but pose significant challenges for real-world adaptation due to the complexities of GAN inversion, mapping real images to their corresponding latent representations.

Traditional GAN inversion methods face a critical trade-off between distortion and editability. Approaches prioritising editability often compromise reconstruction fidelity, failing to capture fine image details, while fidelity-focused methods yield latent representations that resist semantic manipulation. As noted by Katsumata et al [17], this trade-off reflects the inherent difficulty of achieving both faithful reconstructions and flexible editing within the latent space. Generator-tuning approaches, such as those proposed by Roich et al. [18], address this challenge by fine-tuning the generator to embed input images into well-behaved latent regions. While effective, this method is computationally intensive, requiring extensive per-image optimisation.

HyperStyle, introduced by Alaluf et al. [19], offers a transformative solution by integrating generator-tuning techniques within an encoder-based framework. It employs a lightweight hypernetwork to dynamically refine generator weights, aligning them with the input image's unique attributes. The hypernetwork consists of a ResNet-based feature extractor and refinement blocks that predict offsets for StyleGAN's convolutional filters. These offsets enable high reconstruction fidelity while preserving latent space editability. Alaluf et al. highlights design innovations such as parameter sharing, weight sharing, and depthwise-inspired convolutions, which reduce complexity and improve efficiency. HyperStyle also incorporates iterative refinement, progressively optimising generator weights without requiring per-image adjustments, achieving state-of-the-art quality with unmatched scalability.

Building on HyperStyle's robust inversion capabilities, InterfaceGAN enables precise attribute control by leveraging the semantic structure of GAN latent spaces. As detailed by Shen et al. [20], it identifies linear hyperplanes corresponding to interpretable attributes such as pose, age, and gender. These hyperplanes act as decision boundaries, separating latent codes linked to different attribute values. For pose manipulation, a binary classifier trained on StyleGAN-generated images defines a hyperplane that allows controlled adjustments by traversing the latent code along its normal vector.

InterfaceGAN's strength lies in disentangling pose from unrelated attributes by ensuring orthogonality between the pose hyperplane and those representing features such as gender or age. This disentanglement preserves identity and facial characteristics during edits. Additionally, by operating directly in the latent space, InterfaceGAN avoids individualised generator fine-tuning, making it efficient for large-scale applications and high-throughput pipelines.

Together, HyperStyle and InterfaceGAN form a synergistic duo that enhances SCHIGAND's capabilities. HyperStyle addresses the inversion challenges by achieving high reconstruction fidelity while preserving editability, and InterfaceGAN introduces controlled pose variations. This seamless integration enables SCHIGAND to generate diverse, realistic identities tailored to specific needs, advancing synthetic data generation and biometric research.

### 3.3 Diffusion (D)

In SCHIGAND, diffusion models serve as the final, pivotal component, utilising the DreamBooth technique to generate synthetic identities with exceptional fidelity and versatility. As noted by Chen et al., diffusion models have revolutionised image generation by iteratively refining random noise into coherent, high-quality outputs [21]. This process allows for a balanced synthesis of global structure and fine details, particularly when conditioned on textual descriptions, making them ideal for generating realistic, personalised images. SCHIGAND capitalises on these strengths to create synthetic identities tailored to diverse requirements and applications.

The DreamBooth technique, introduced by Ruiz et al., fine-tunes pre-trained text-to-image models using only a small set of images [22]. It incorporates class-specific prior preservation loss to embed a subject's unique identity into the model's semantic space while retaining generalisability. By requiring as few as three to five images, this approach enables the diffusion model to faithfully represent the subject across varied poses, environments, and contexts without extensive datasets, democratising access to high-quality, personalised image synthesis.

SCHIGAND initially employed Stable Diffusion version 2 (SD v2) for its robust text-to-image generation capabilities and computational efficiency. SD v2 builds on the latent diffusion framework introduced by Rombach et al., compressing latent spaces to produce high-fidelity images while reducing computational demands [23]. This efficiency provided a strong baseline for photorealistic identity synthesis. To achieve higher-resolution outputs and finer detail, SCHIGAND later integrated Stable Diffusion XL (SDXL), which features an enhanced architecture, including a larger UNet backbone and dual text encoders, significantly increasing its parameter count and capacity for detailed image synthesis [24].

However, SDXL's increased complexity posed computational challenges specifically large memory requirements, which SCHIGAND addressed using Low-Rank Adaptation (LoRA). LoRA efficiently updates only a small subset of the model's parameters by introducing trainable low-rank matrices, significantly reducing memory requirements while retaining performance [25][26]. This combination of LoRA and SDXL enabled SCHIGAND to achieve both high-resolution output and computational efficiency, enhancing its ability to generate superior identity representations.

The fusion of InterfaceGAN's latent space manipulations with the diffusion model's text-to-image generation capabilities represents a major innovation in identity synthesis. InterfaceGAN ensures a diverse and robust initial dataset, while DreamBooth fine-tuning imbues the diffusion model with the ability to preserve identity fidelity across various contexts. This scalable and efficient integration positions SCHIGAND as a transformative tool for synthetic data generation.

By balancing fidelity and flexibility, SCHIGAND unlocks new possibilities in biometric research, identity-based testing, and realistic image simulations.

# 4 Evaluation Methodology

In this section, we conduct quantitative tests that compare our proposed SCHIGAND pipeline with existing models and authentic data.

## 4.1 Facial Verification

We compare images using a pre-trained facial biometric model. ArcFace was chosen for its proven effectiveness in biometric research and its ability to handle "in-the-wild" datasets with variations in pose, lighting, and expression. This makes it well-suited for evaluating intra-class variation, inter-class variation, and identity preservation. Its widespread use in research, including the GANDiffFace study, also ensures comparability with existing findings.

## 4.2 Dataset Selection

The VGGFace2 dataset was chosen to evaluate inter-class variation, intra-class variation, and identity preservation due to its extensive intra-class diversity and widespread use as a benchmark [27]. Its varied images shown in Figure 5 make it ideal for comparison with SCHIGAND, where generating variations within an identity is more feasible than creating new identities. Additionally, VGGFace2's role in the GANDiffFace study ensures comparability with existing research. To compensate for its lack of identity labels and demographic annotations, tools like Google Lens were used to extract attributes such as ethnicity, enabling a more comprehensive biometric analysis.

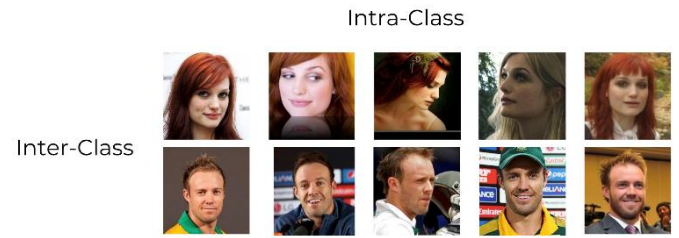

*Figure 5. Example images from the VGGFace2 dataset showing five images for two separate identities.*

## 4.3 Intra-class and Inter-class Variation

The methodology for evaluating intra-class and inter-class variation follows the approach used in GANDiffFace, enabling direct comparisons between synthetic datasets and real-world benchmarks such as VGGFace2. This involves computing similarity scores for mated and non-mated pairs using feature embeddings extracted by ArcFace.

1. *Similarity Scores and Pairings*

For each identity in a dataset, $N = 10$ image variations are selected and paired as follows:

- Intra-Class Variation (Mated Pairs): 20 image pairs are randomly selected from all possible combinations of the 10 variations of the same identity. These pairs assess variations in pose, expression, and other intra-identity features.

- Inter-Class Variation (Non-Mated Pairs): 20 random pairs are formed between images of

different identities to measure the distinctiveness of identities.

Cosine similarity is used to compute the similarity score for each pair as defined in Equation 1. Here $f_i$ and $f_j$ are the feature embeddings of images $i$ and $j$, and $||f||$ represents the Euclidean norm. Scores are normalised to the range [-1, +1], with higher values indicating greater similarity.

$$S(i,j) = \frac{f_i \cdot f_j}{||f_i|| \; ||f_j||}, \quad S(i,j) \in [-1, +1] \quad (1)$$

The mean and standard deviation of the mated ($S_m$) and non-mated ($S_{nm}$) scores summarise intra-class and inter-class variations, respectively. These are calculated as shown in Equations 2, 3, and 4.

$$\mu_m = \frac{1}{N_m} \sum_{k=1}^{N_m} S_m(k), \; \mu_{nm} = \frac{1}{N_{nm}} \sum_{l=1}^{N_{nm}} S_{nm}(l) \quad (2)$$

$$\sigma_m = \sqrt{\frac{1}{N_m} \sum_{k=1}^{N_m} (S_m(k) - \mu_m)^2} \quad (3)$$

$$\sigma_{nm} = \sqrt{\frac{1}{N_{nm}} \sum_{l=1}^{N_{nm}} (S_{nm}(l) - \mu_{nm})^2} \quad (4)$$

where $N_m$ and $N_{nm}$ denote the number of mated and non-mated comparisons respectively. A higher $\mu_m$ indicates better identity preservation, while a lower $\mu_{nm}$ reflects greater separation between identities. Achieving high intra-class diversity ($\sigma_m$) while maintaining low inter-class similarity ($\mu_{nm}$) is crucial for generating realistic and effective synthetic datasets. By following this methodology, synthetic datasets are benchmarked against VGGFace2 and GANDiffFace to assess their suitability for biometric applications.

## 2. KL (Kullback-Leibler) Divergence

KL Divergence evaluates how closely the non-mated or mated score distributions of a synthetic dataset align with those of a real dataset by capturing both mean and standard deviation within a single metric. Unlike mean and standard deviation, which summarise specific aspects of the distributions, KL Divergence provides a complete measure of the difference between the probability distributions. It captures subtle variations in the shape, skewness, and modality of the distributions, offering a more detailed comparison. Additionally, its use in

GANDiffFace establishes it as a standard benchmark for comparative evaluation.

To compute KL Divergence, cosine similarities ranging from [−1,1] were standardised to ensure compatibility with histogram-based probability distributions [28] while maintaining their interpretability.

This standardisation was achieved by adding one to the cosine distance and dividing the result by two to obtain a normalised cosine distance in the range [0,1]. Similarity was then computed by subtracting this normalised value from one. This ensures cosine distances are suitable for constructing probability histograms.

Once transformed, the cosine distances were binned into discrete intervals to create histograms for both the real ($P$) and synthetic datasets ($Q$). These histograms were normalised to represent valid probability distributions as shown in Equations 5 and 6.

$$P(x) \leftarrow \frac{P(x)}{\sum_x P(x)} \quad (5)$$

$$Q(x) \leftarrow \frac{Q(x)}{\sum_x Q(x)} \quad (6)$$

The KL Divergence between the distributions was calculated using Equation 7.

$$KL(P|Q) = \sum_x P(x) \ln\left(\frac{P(x)}{Q(x)}\right) \quad (7)$$

To address issues with division by zero, a small constant $\epsilon = 10^{-10}$ is added to each bin as shown in Equations 8 and 9.

$$P(x) \leftarrow P(x) + \epsilon \quad (8)$$

$$Q(x) \leftarrow Q(x) + \epsilon \quad (9)$$

This ensured numerical stability during calculation. The implementation leveraged the *rel_entr(P, Q)* function from *scipy.special*, which computes the relative entropy for each bin efficiently.

The decision to normalise cosine distances using $(\frac{x+1}{2})$ ensures consistent mapping to the range ([0, 1]) across all datasets. This method works well with histogram-based probability distributions and avoids issues with outliers affecting the range and resolution, a case seen in SCHIGAND and VGGFace2 when normalising by the maximum cosine similarity.

### 4.4 Identity Preservation

While the means of both distributions in the previous section can be used as an indicator to identity preservation, it can be insufficient since variability is ignored within the distributions. Although the mean is a useful indicator, it does not capture deviation. For instance, a low mean combined with a high standard deviation may incorrectly suggest poor identity preservation, when the real challenge lies in the model's ability to handle dataset diversity.

KL Divergence offers an alternative by comparing the overall alignment of score distributions. Matching distributions would indicate similar identity preservation while accounting for both mean and standard deviation. However, the inherent variability of "in-the-wild" datasets like VGGFace2 undermines the validity of such comparisons as a baseline. Additionally, distribution-based measures assume that the biometric model can perfectly distinguish identity variations, which rarely holds true in practice.

To address these limitations, a verification threshold calculated using the Equal Error Rate (EER) provides an independent measure of performance. By evaluating the system's ability to separate mated and non-mated pairs while accounting for overlap in their distributions, the EER reflects real-world decision-making capabilities.

To evaluate verification performance, we computed the Equal Error Rate (EER), which represents the point at which the False Acceptance Rate (FAR) equals the False Rejection Rate (FRR). This threshold-independent metric provides an assessment of overlap between mated and non-mated similarity score distributions. Standard values such as the True Positive Rate (TPR) and False Positive Rate (FPR) were computed using the $roc\_curve$ $()$ function from $scikit-learn$, and the EER threshold $\tau_{EER}$ was identified by minimising the absolute difference between FAR and FRR. A small constant $\epsilon = 10^{-10}$ was added to the resulting FPR and FRR values to ensure numerical stability.

### 4.5 Synthetic Dataset Generation Strategy

A key decision was determining which aspects to control in our synthetic dataset: demographics, environmental variations, or both.

As a result, we chose to control ethnicity and gender for each identity while selecting a sample of random variations. This approach also facilitated the investigation of demographic bias between data sets, particularly in the preservation of identity.

To ensure that the synthetic datasets, DCFace and SCHIGAND, contained "in-the-wild" variations comparable to those in VGGFace2, specific strategies were implemented. For DCFace, the style bank, as detailed in Section 2, was designed to include a diverse range of ethnicities and poses. For SCHIGAND, prompts were carefully curated to maximise within-identity variation by altering attributes such as pose, lighting, expression, and background.

Ethnicity selection was guided by the classifications proposed by Deng et al., which included: Asian, Black, White, Middle Eastern, Hispanic, and Indian [29]. However, testing with SCHIGAND revealed issues with these categories. For instance, specifying "Black" or "White" often resulted in unnaturally dark or light images, while the term "Asian" was too ambiguous, encompassing both East Asian and South Asian features. Consequently, the refined ethnicity categories used were: East Asian, African, Caucasian, Middle Eastern, Hispanic, and Indian.

Based on these considerations, we determined that 36 identities would suffice for a proof-of-concept trial, representing six ethnicities, with each split into three male and three female identities. Each selected identity contained 70 variations, ensuring a large pool of images for the random selection of mated and non-mated scores.

## 5   Results and Analysis

This section presents the quantitative analysis of the generated synthetic datasets, based on the evaluation methodology described previously. The performance of the DCFace and SCHIGAND models is compared against the real-world VGGFace2 benchmark. The results are examined through two angles: first, the trade-off between generating intra-class variations versus creating distinct inter-class separation; and second, how well the models can preserve identity between creating variations of the same person.

### 5.1 The Trade-Off: Simulating "In-the-Wild" Variation vs. Identity Separation

Our analysis reveals a distinct trade-off between the two synthetic generation methods, highlighting their suitability for different biometric testing applications. The data shows that while SCHIGAND excels at generating realistic intra-class variation, DCFace is more effective at creating clearly separable inter-class identities.

- Intra-Class Variation (Diversity): SCHIGAND demonstrates a superior ability to generate diverse,

"in-the-wild" variations for a single identity, closely aligning with that of the VGGFace2 dataset. As shown in Table 1, the standard deviation of its mated scores (0.21) is significantly higher than DCFace's (0.11) and is comparable to the real VGGFace2 benchmark (0.18). This indicates a more realistic spread of appearances for each identity. This finding is strongly supported by the Kullback-Leibler (KL) Divergence results in Table 2 where SCHIGAND's mated score distribution achieves a much lower (i.e., better) divergence score of 0.17 compared to DCFace's 1.87. The similarity in distribution shape between SCHIGAND and VGGFace2 is visually evident in Figure 6.

- Inter-Class Variation (Distinctiveness): However, DCFace is more effective at producing clearly distinct identities that are easily separated by the verification model. Table 1 shows its mean non-mated score $(0.17 \pm 0.10)$ is lower than SCHIGAND's $(0.28 \pm 0.18)$, indicating less similarity between different individuals. The KL Divergence scores in Table 2 confirm this advantage, with DCFace achieving a non-mated divergence of 0.37, which is less than half of SCHIGAND's 0.85. This suggests that the identities generated by DCFace are more clearly defined and separated, as visualised in Figure 7.

| Dataset | Type | Id. | Mated Scores | Non-Mated Scores | Backbone | Embeddings |
|---|---|---|---|---|---|---|
| GAN-based | Syn | 700 | 0.67 ± 0.14 | 0.08 ± 0.10 | iresnet100 | ArcFace |
| GANDiffFace (tip=0.4) | Syn | 700 | 0.59 ± 0.12 | 0.08 ± 0.09 | iresnet100 | ArcFace |
| GANDiffFace (tip=0.3) | Syn | 700 | 0.55 ± 0.15 | 0.08 ± 0.09 | iresnet100 | ArcFace |
| GANDiffFace (tip=0.2) | Syn | 700 | 0.51 ± 0.17 | 0.07 ± 0.09 | iresnet100 | ArcFace |
| SFace | Syn | 411 | 0.18 ± 0.13 | 0.02 ± 0.08 | iresnet100 | ArcFace |
| DigiFace-1M | Syn | 2,000 | 0.47 ± 0.15 | 0.12 ± 0.09 | iresnet100 | ArcFace |
| VGGFace2 | Real | 8,515 | 0.52 ± 0.16 | 0.01 ± 0.07 | iresnet100 | ArcFace |
| IJB-C | Real | 2,557 | 0.57 ± 0.17 | 0.01 ± 0.07 | iresnet100 | ArcFace |
| DCFace | Syn | 36 | 0.52 ± 0.11 | 0.17 ± 0.10 | retinaface | ArcFace |
| SCHIGAND | Syn | 36 | 0.56 ± 0.21 | 0.28 ± 0.18 | retinaface | ArcFace |
| VGGFace2 | Real | 36 | 0.54 ± 0.18 | 0.12 ± 0.12 | retinaface | ArcFace |

*Table 1. Comparison of mated and non-mated scores across synthetic and real datasets. Mated scores reflect identity preservation (intra-class similarity), while non-mated scores reflect inter-class variability. Metrics are computed using a RetinaFace backbone.*

Overall Insight: This analysis highlights a critical trade-off. For practitioners seeking to evaluate a biometric system's robustness against challenging, high-variation conditions, SCHIGAND provides a more realistic and suitable dataset. For applications where the goal is to generate a set of highly distinct identities with less intra-class complexity (e.g., for enrolment or simple portrait-based testing), DCFace is the more effective tool.

| Dataset | Mated Scores | | Non-mated Scores | | EER | Backbone | Embeddings |
|---|---|---|---|---|---|---|---|
| | VGG2 | IJB-C | VGG2 | IJB-C | | | |
| GAN-based | 0.69 | 0.28 | 0.48 | 0.42 | 1.49% | iresnet100 | ArcFace |
| GANDiffFace (tip=0.4) | 0.16 | 0.09 | 0.52 | 0.46 | 1.25% | iresnet100 | ArcFace |
| GANDiffFace (tip=0.3) | 0.16 | 0.16 | 0.48 | 0.43 | 2.74% | iresnet100 | ArcFace |
| GANDiffFace (tip=0.2) | 0.23 | 0.28 | 0.42 | 0.37 | 5.11% | iresnet100 | ArcFace |
| SFace | 1.72 | 2.11 | 0.18 | 0.11 | 22.53% | iresnet100 | ArcFace |
| DigiFace-1M | 0.21 | 0.41 | 1.05 | 1.02 | 7.92% | iresnet100 | ArcFace |
| VGGFace2 | 0.11 | - | 0.01 | - | 4.51% | iresnet100 | ArcFace |
| IJB-C | 0.15 | - | 0.01 | - | 3.22% | iresnet100 | ArcFace |
| DCFace | 1.87 | - | **0.37** | - | 5.69% | retinaface | ArcFace |
| SCHIGAND | **0.17** | - | 0.85 | - | 21.25% | retinaface | ArcFace |
| VGGFace2 | - | - | - | - | 8.61% | retinaface | ArcFace |

*Table 2. Comparison of mated/non-mated scores, KL divergence, and EER across datasets. Lower KL and EER indicate better identity preservation. Metrics computed using ArcFace embeddings with IRSEnet100 or RetinaFace backbones.*

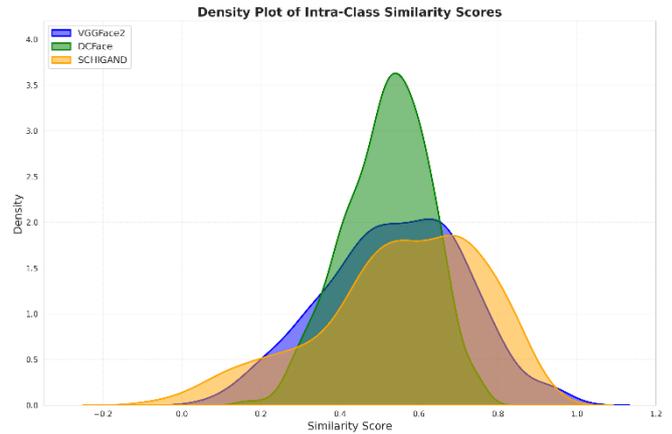

*Figure 6. Density plot of intra-class (mated) similarity scores. The distribution for SCHIGAND more closely resembles the shape and spread of the real VGGFace2 data compared to the narrow distribution of DCFace.*

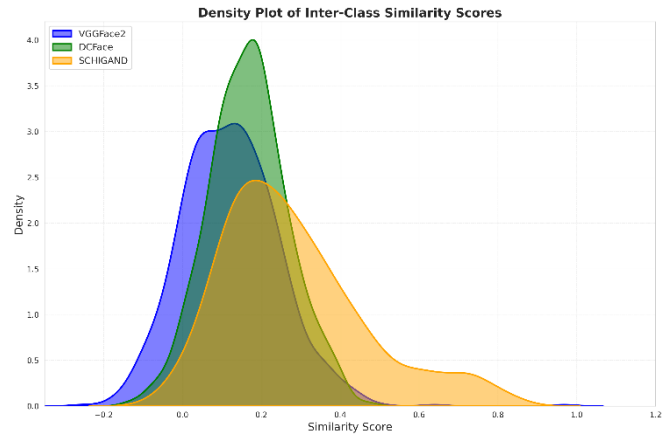

*Figure 7. Density plot of inter-class (non-mated) similarity scores, representing the similarity between different classes. The DCFace distribution shows a tighter cluster at lower similarity values, indicating better identity separation.*

Accurate identity preservation is essential for a synthetic dataset to be useful. This section evaluates how well the identities generated by SCHIGAND and DCFace are preserved across different samples, using standard verification metrics to quantify the difficulty of this task. A dataset that is challenging yet consistent provides a more robust benchmark.

Verification Results: The Equal Error Rate (EER), shown in <u>Table 3</u>, measures the point where the rates of false acceptances and false rejections are equal. On this metric, DCFace achieves the lowest EER of 5.69%, indicating that identities within its dataset are the easiest to distinguish and verify correctly. The real VGGFace2 subset presents a more challenging task (8.61% EER), while the SCHIGAND dataset is the most challenging, with the highest EER of 21.25%. The ROC curve in <u>Figure 8</u> visually confirms this, with DCFace's curve demonstrating the highest performance.

Analysis of Identity Preservation: These verification results are a direct consequence of the dataset characteristics analysed in <u>Section 5.1</u>.

- DCFace: Its strong performance (low EER, high TPR of 0.9458) stems from its low intra-class variation. The generated images for each identity are highly consistent, which makes the task of preserving identity straightforward for a verification model.

- SCHIGAND: Its high EER (21.25%) and lower TPR (0.7889) signify that preserving identity across its samples is a much more difficult task. This is caused by the high, realistic "in-the-wild" variation deliberately generated by the pipeline. The significant changes in pose, lighting, and expression make it harder for the verifier to consistently match two images of the same person.

Overall Insight: This analysis reveals that while DCFace produces identities that are more easily preserved by a verifier, this is due to its limited variation. SCHIGAND, in contrast, creates a more difficult and realistic identity preservation challenge. Therefore, while DCFace's EER score is better on paper, SCHIGAND serves as a more effective benchmark for assessing a verification system's ability to handle the identity preservation challenges found in real-world data.

### 5.3 Qualitative Analysis

Following the quantitative analysis seen in <u>Figure 9</u>, selected images from each model are showcased to highlight their different visual characteristics. For example, within the selected DCFace images, all the outputs appear to be from the same frontal angle. Comparatively, the SCHIGAND and VGGFace2 outputs represent a much more varied angle and pose for each identity.

| Model | EER (%) | Threshold at EER | TPR at EER |
|---|---|---|---|
| SCHIGAND | 21.25 | 0.4087 | 0.7889 |
| VGGFace2 | 8.61 | 0.2820 | 0.9139 |
| DCFace | 5.69 | 0.3351 | 0.9458 |

*Table 3. Comparison of EER, threshold at EER, and TPR at EER for each dataset. Lower EER indicates better distinction between same and different identities, reflecting stronger identity preservation.*

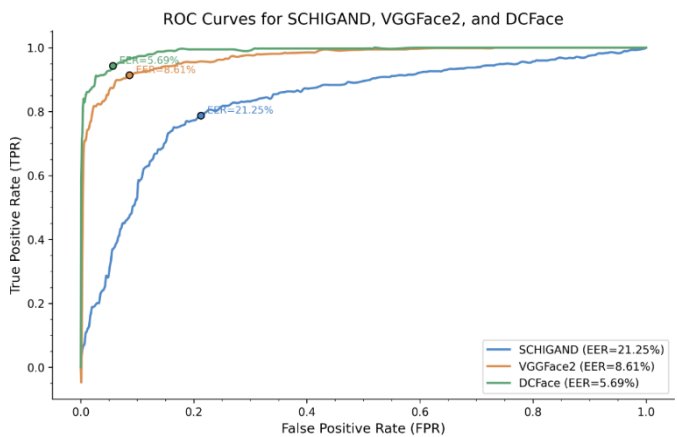

*Figure 8. ROC curves comparing verification performance on SCHIGAND, VGGFace2, and DCFace. DCFace shows stronger identity preservation, while SCHIGAND's curve reflects increased variation and verification difficulty.*

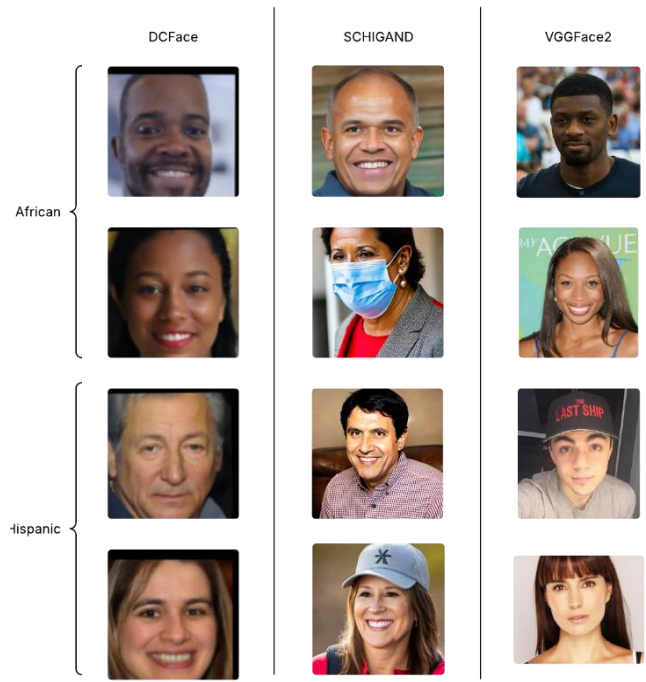

*Figure 9. Comparison of synthetically generated images (by DCFace and SCHIGAND) to the real dataset - VGGFace2.*

## 6    Limitations

Although the experiment and data structure detailed above are comprehensive, they have been restricted and adjusted to suit the allocated resources and time constraints. In addition to the limitations mentioned throughout this section, the following are additional ones that should be considered:

- The ArcFace model's backbone used for testing GANdiffFace, iResNet100, could not be utilised due to compatibility issues with the Open Neural Network Exchange (ONNX) runtime (a high-performance engine for executing ML models) on Windows Subsystem for Linux (WSL). As a substitute, RetinaFace was employed, a backbone explicitly designed for applications in uncontrolled environments, as outlined in Deng et al. [30].

- While our quantitative analysis offers broad coverage across key performance areas, it does not fully capture perceptual realism as interpreted by humans. Models such as ArcFace focus on geometric features and identity-related embeddings but may overlook visual artefacts or subtle details that influence perceived authenticity. Incorporating qualitative evaluation alongside quantitative methods could provide a more comprehensive assessment.

- The limitation of time and computational resources has reduced the size of the synthetic datasets generated. However, the information can still serve as a foundation for further research and exploration.

## 7    Conclusion

This research successfully met its objectives by investigating and implementing state-of-the-art GenAI models for synthetic dataset generation in facial biometrics. We conducted a comprehensive review of state-of-the-art GenAI methodologies for synthetic facial imagery, evaluating several models and excluding some based on performance limitations or ethical concerns. This review was instrumental in the development of SCHIGAND, a novel pipeline designed to enhance synthetic dataset generation.

We successfully implemented DCFace and SCHIGAND to generate synthetic facial datasets. DCFace provided strong demographic representation but lacked environmental context, while SCHIGAND offered greater diversity in both identity generation and environmental conditions.

Both models were evaluated against real datasets using biometric metrics. DCFace exhibited high inter-class variation, whereas SCHIGAND demonstrated superior intra-class diversity. Identity preservation was assessed with ArcFace, confirming that these models can potentially supplement real-world data for biometric applications. These findings highlight the potential of synthetic datasets to address key challenges in facial biometrics, offering new opportunities for future research and development.

## 8    Future Work

The following key considerations for future work have been established:

- Completely training facial verification models using synthetic data generated by SCHIGAND could present an opportunity for deeper validation and improved accuracy. This could help gauge its outputs as a replacement or supplement for real facial samples.

- Conduct a qualitative survey on the fidelity of synthetically generated faces to inform further refinements to the SCHIGAND pipeline.

- A dedicated exploration of ethical considerations for the responsible use of synthetic image generation techniques. This focus would strengthen the ethical framework for generating datasets with SCHIGAND and similar models, ensuring compliance with societal and legal standards.

age," in 2018 13th IEEE international conference on automatic face & gesture recognition (FG 2018), pp. 67–74, IEEE, 2018.